# The Association of Transformer-based Sentiment Analysis with Symptom Distress and Deterioration in Routine Psychotherapy Care

**Douglas K. Faust[1,2*], Peter Awad[1], Alexandre Vaz[1], Tony Rousmaniere[1]**

[1]Sentio University, Torrance, CA, USA

[2]Western Washington University, Department of Mathematics, Bellingham, WA, USA

*** Correspondence:**
Douglas K. Faust
faustd@wwu.edu

## Abstract

Sentiment analysis has been of long-standing interest in psychotherapy research. Recently, the Transformer deep learning architecture has produced text-based sentiment analysis models that are highly accurate and context-aware. These models have been explored as proxies for emotion measurement instruments in psychotherapy, but not investigated as stand-alone psychometric tools. Using proposed utterance-level and session-level sentiment features derived from a fine-grained sentiment model on a large corpus of psychotherapy sessions (N = 751), we investigate the distribution of session aggregated sentiment scores. Further, we characterize the relationship of these features to individual components and the overall score of the OQ-45 instrument and find that this sentiment feature is most strongly correlated to components related to emotional valence in directionally intuitive ways. Finally, we report that there are statistically significant differences between the sentiment distributions for patients flagged as at risk of deterioration or dropping out of care via either the OQ Rational or Empirical outcome models. These correlations to a fully-validated psychometric instrument demonstrate that these proposed sentiment features are, at least, adjunctive measures of client distress and deterioration.

## 1 Introduction

Sentiment analysis has long been of interest in psychotherapy research owing to the large role that emotions are understood to play in the therapeutic processes [1]. Researchers or therapists may subjectively assess clients' perceived sentiment or clients may self-report emotional states. Sentiment scores can also be obtained from session recordings by machine learning (ML) models to produce objective metrics. Although assessing emotion or sentiment is inherently multimodal and considers words, speech cadence and prosody, facial expressions [2] and gestures [3], text-based measurements are the most well-studied and reliable. Recently, fast and accurate Transformer models trained on massive amounts of labeled text [4] have triggered a resurgence in interest in sentiment analysis of psychotherapy session text.

Many natural language processing (NLP) techniques have been used to identify emotion in text see [5] for a review. The Transformer is a form of deep neural net with an "attention" mechanism [6] and NLP models built on the Transformer architecture are a significant breakthrough in text analysis. Previously, text mining to identify emotions was largely done with "Lexicon" methods using dictionaries of negative and positive emotion words [7, 8]. Lexicon methods are simple but severely limited as each word is considered and scored individually. Negation, for example, which can completely change the emotional polarity of a statement cannot be captured at the single-word level. In contrast, Transformer-based sentiment models (TSM) are trained on entire sections of text like product reviews or social media posts and report sentiment scores at the utterance, not word, level.

Eberhardt et al. [9] recently used a two-category (positive, negative) TSM model and correlated session text sentiment to the therapist and client's reported emotional valence via the "Profile of Mood States" instrument [10]. They reported high correlations between TSM predictions and the self-scored and therapist-scored negative emotion.

Atzil-Slonim et al [11] also recently used a TSM to investigate the coherence between clients' self-reported emotional state and their expressed in-session sentiment. Further, they showed that increased coherence in between the measured and self-reported negative sentiment was associated with improvement in functioning, validating theory expressed in [12] and [13] that emotional coherence is associated with better functioning.

Because of the quality of TSM, they should also be investigated as stand-alone psychometric tools as well as instrument proxies, as reported above. Direct measurements of sentiment from therapy sessions are appealing since they do not have compliance or completeness issues versus a separately-administered instrument. Furthermore, these data are objective and thus offer a complementary perspective to self-reported values. To date, ML has already proven useful across many domains of mental health research [14] (e.g., mental health detection: Abd-Alrazaq et al. [15]; mobile mental health: Goldberg et al. [16]; emotion measurement Tanana et al.[17]). For example, Atzil-Slonim et al. [18] used topic models to identify clients' levels of functioning and alliance ruptures, and Smink et al. [19] employed text mining to relate change processes to therapeutic outcomes.

In this work, we propose a sentiment score feature describing entire psychotherapy sessions and report the statistics of this feature over 751 recorded telehealth therapy sessions. Before the utterance-level scores are averaged to produce a final session score, this method also produces a time-series of how sentiment evolves over individual sessions, providing a rich space for future investigations of in-session sentiment dynamics.

The Outcome Questionnaire 45 (OQ-45) is used in this study as the benchmark for evaluating the clinical relevance of TSM scores. The OQ-45 is a widely used, psychometrically robust, patient reported outcome measure that captures global psychological distress across symptom severity, interpersonal functioning, and social role performance. These domains have been shown to align closely with the emotional experiences that unfold in psychotherapy sessions [20]. Because it is extensively normed and provides established cutoff scores and indices of reliable change, the OQ-45 offers a clinically interpretable standard for determining whether automatically derived sentiment scores map onto meaningful variation in patients' mental health status. This instrument's routine use in many clinical settings further supports its suitability as a reference point for validating new automated assessment methods [21].

In this study, linking session level sentiment scores to OQ-45 data provides a test of both concurrent and longitudinal validity. The progress and alert categories of the OQ-45 are calibrated to indicate improvements or deteriorations in clients' reported distress [22] and the correlation of a sentiment model feature with these categories would show that expressed verbal sentiment can indicate progress or lack of progress in treatment.

The study also highlights practical considerations for conducting clinical psychological research. The data pipeline demonstrates the use of automated transcription, diarization, and speaker role attribution to substantially reduce the resource burden associated with processing large volumes of session data. The TSM used here was also deployed locally on commodity hardware to maintain strict confidentiality of sensitive clinical material, thereby avoiding the privacy risks inherent in cloud-based data and computing services.

# 2 Materials and Methods

## 2.1 Clinical Setting

### 2.1.1 Therapists

There were 11 therapists included in the study. Nine were graduate students in masters marriage and family therapy programs who had no previous clinical experience, and two were associate marriage and family therapists with one year of previous clinical experience. Therapists' ages ranged between 24 and 59; five were male, and six were female. Regarding race/ethnicity, 8 therapists were White/European (72.7%), one Asian American/ Pacific Islander (9.0%), one Middle Eastern (9.0%), and one Black/African American (9.0%). In the data set used in this study, trainees each saw an average of 6.9 clients (SD = 4.1, Mdn = 6.5).

### 2.1.2 Clients

There were in total 94 clients included in the study, who were seen by the aforementioned 11 therapists. Among these clients, 52 identified as cisgender female (55.1%), 31 identified as cisgender male (32.6%), two identified as transgender (2.4%), one reported other (0.8%), and nine provided no responses (9.1%). Regarding race/ethnicity, 30 clients were White/European American (32.1%), 15 Asian American/Pacific Islander (16.3%), 14 Black/African American (14.7%), 11 multiracial (12.2%), 20 Latinx (21.4%), one Middle Eastern (1.2%), and two other (2.1%). Client ages were between 18 and 79. For these clients, the most common complaints were anxiety, depression, posttraumatic stress disorder, and adjustment disorder.

### 2.1.3 Treatment

Counseling was conducted at Sentio Counseling Center, the practicum site of the Sentio University Master of Arts in Marriage and Family Therapy program, between 2023 and 2025. The Sentio Counseling Center offers low-fee online therapy services for adults in the US state of California. At this center, client outcome data are collected continuously throughout treatment as part of routine quality improvement and supervision practices. Sessions lasted 50 minutes and were generally held once per week. On average, clients attended 10 sessions (SD = 7, range: 2–32). Before beginning therapy, all clients provided informed consent agreeing to participate in treatment and allow their anonymized data to be used for supervision and research purposes. Data from 751 therapy sessions were included in this study.

### 2.1.4 Measures

Outcome Questionnaire (OQ-45.2). The OQ-45.2 [23] is a well-established self-report instrument used to assess overall psychological functioning. Because it is sensitive to short-term changes in distress, it can be administered weekly to track clinical progress [23]. Respondents rate the frequency of psychological symptoms experienced over the past week using a 5-point scale from 0 ("never") to 4 ("almost always"), with higher scores reflecting greater distress. The measure includes three subscales—Symptom Distress (e.g., "I feel blue"), Interpersonal Relations (e.g., "I feel lonely"), and Social Role (e.g., "I feel stressed at work/school")—which combine to yield a total score between 0 and 180. A total score of 64 or higher indicates clinically significant distress, while a change of 14 or more points represents a reliable change beyond expected measurement error. The OQ-45.2 demonstrates strong psychometric properties, including internal consistency ($\alpha = .93$) and test–retest reliability ($r = .84$) for the total score [23, 24]. In this dataset, internal consistency ranged from .77 to .84.

#### 2.1.5 Training Program

Therapists participated in three hours of deliberate practice–based supervision weekly, following the Sentio supervision model [25]. Supervision was provided by licensed clinicians who held a master's or doctoral degree, at least five years of experience, and advanced training in deliberate practice supervision. All supervision sessions were video-recorded, and supervisors met weekly for "supervision-of-supervision" meetings with the clinic's training director to review footage and maintain consistency. The approach reflects growing support for experiential learning in therapist education (e.g. [26, 27]) and evidence from numerous recent studies showing deliberate practice training produces stronger clinical skill acquisition than conventional instruction [28, 29]. In addition to supervision, trainees attended a two-hour weekly clinical seminar covering major treatment models, including cognitive behavioral therapy, motivational interviewing, emotion-focused therapy, and schema therapy.

Routine outcome monitoring (ROM) is a standard part of this training model. Therapists are required to review the client's OQ-45.2 results before each session and are trained in interpreting scores, managing in-session risk, documenting appropriately, and integrating feedback into treatment. The program provides clear protocols and resources for discussing OQ data with clients.

### 2.2 Data Collection

#### 2.2.1 Survey Administration

Clients completed the OQ-45.2 online before each session through a secure portal. Therapists accessed client reports through the same portal, which automatically logged user IDs for data security. The scheduling system also recorded time stamps for all sessions. Prior to analysis, identifying details were removed and replaced with anonymous ID codes.

#### 2.2.2 Data Pipeline

The Sentio Counseling Center clinic is a fully remote telehealth provider which uses a HIPAA compliant video conferencing platform to host and store session recordings and notes. Within a day of recording, the session videos and their recording metadata are moved to a local HIPAA-compliant compute and storage environment and matched to the session notes and OQ survey results associated with their particular session. Recordings shorter than 45 minutes were excluded from the study. This length exclusion was chosen as the midpoint in between two modes of a bimodal length distribution corresponding to full and interrupted or incomplete sessions.

Therapy sessions were automatically transcribed, diarized, and segmented using a secure local processing pipeline before further human quality validation, providing a privacy-preserving and scalable approach to preparing the data. This process is schematized in Figure 1.

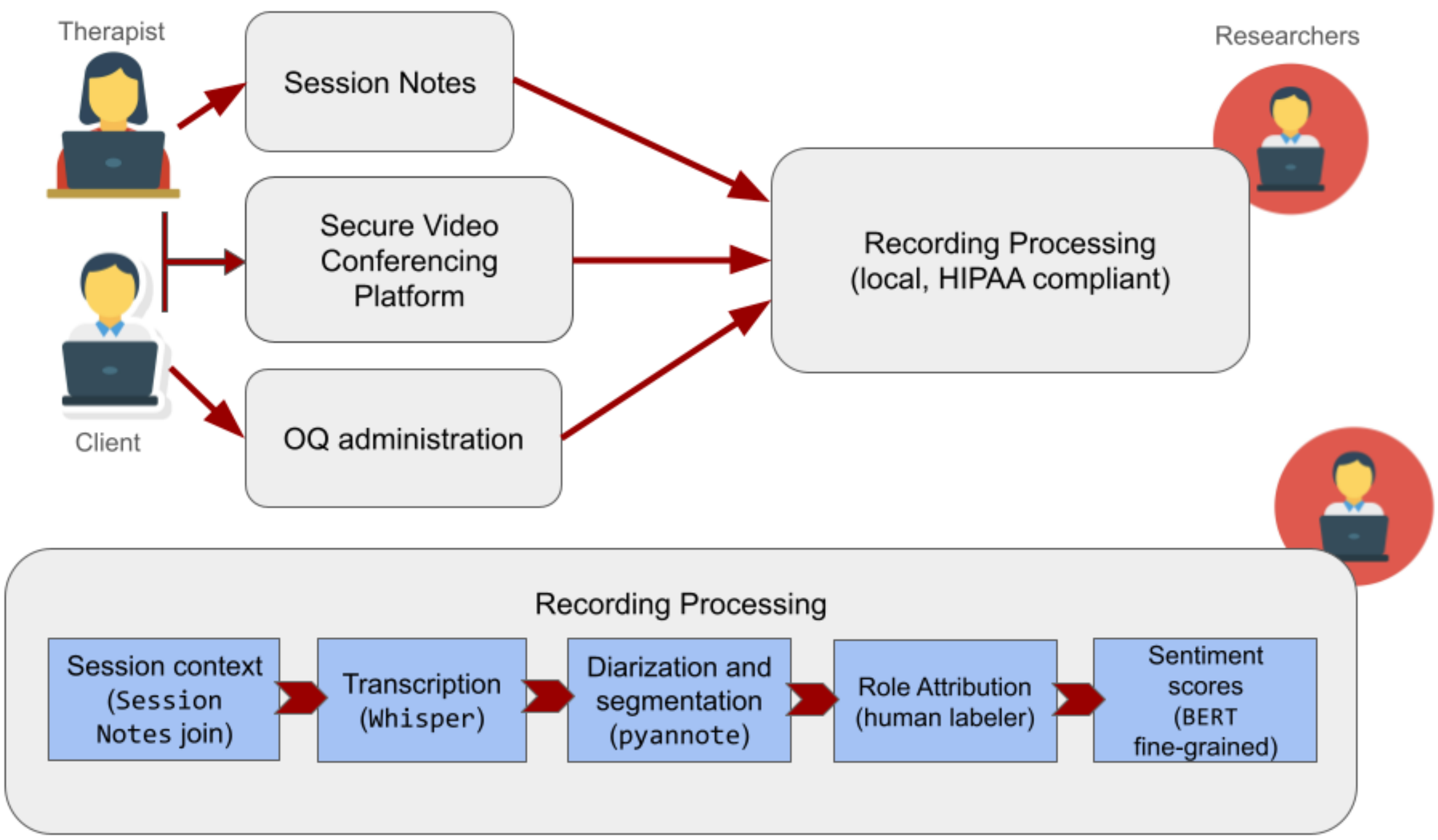

Fig. 1 Overview of the system used to collect and process recorded psychotherapy sessions. The quality control and transcript correction is done at the "Role Attribution" stage of the recording processing.

The proposed transcription was done by WhisperX 'large-v3' model [30] with automatic language detection in "segment" alignment mode.

The proposed diarization was done by the pyannote open-source toolkit written in the Python programming language [31]. Specifically, pyannote-audio code version 3.1.1 [32] with the number of speakers forced to 2 commensurate with the fact that the sessions in this dataset are all individual therapy sessions.

Finally, the transcripts were processed into utterance level segments with both turn-level and word-level timestamps for future, more granular, analysis. The speaker labels assigned to each turn at this point in the data pipeline segregate the turns of generic "SPEAKER_01" and "SPEAKER_02" participants, but do not identify the speaker's role.

The WhisperX and pyannote models, as well as all other processing was run on access-controlled on-premesis computing resources to ensure the privacy of the participants.

The proposed transcript and original video were then sent to be validated and, if need be, corrected by a researcher. This annotator was also responsible for identifying the speaker roles (e.g. SPEAKER_01 = “client”, SPEAKER_02 = “therapist”), but we note that role classification could also have been automated via simple language features like speaker talk time, number of questions asked, and the existence of a standard compliance question that the state of California requires telehealth providers to ask before each session.

A similar data pipeline has recently been employed by Eberhardt et al [33] to create 1,131 transcribed and diarized sessions for the purposes of automated labeling of “engagement” via fine-tuned LLM. The pipeline used in this study differs from that work in a few key respects. The quality validation performed in [33] was done before the transcription and speaker embeddings were used to assist diarization. In this work, the full diarized transcripts were validated and corrected post-processing. When possible, this validation was done by the therapist who participated in the session and otherwise by the clinic director or a researcher who was assigned all of the videos from a particular client-therapist pair for consistency.

In a telehealth clinic, clients may connect from a low quality connection, either due to poor internet connection, external environment disruptions, or low audio/video quality (e.g. the client connected to the session by phone). Because of this, human validation was also used to exclude these recordings from analysis when the proposed transcription or diarization met a threshold number of errors.

## 3 Data Analysis

### 3.1 Utterance Sentiment Value

For the sentiment calculation, we used the “large” variant of the original fine-grained BERT based sentiment model released in 2019 by Munikar et al. [34] which has been reproduced and hosted on the huggingface model hub [35]. As with the data pipeline, we deliberately chose an open sourced model hosted on a large platform to help facilitate continued research by avoiding using or creating proprietary or self-hosted models.

The Munikar et al. fine-grained BERT model was validated on the Stanford Sentiment Treebank datasets two (SST-2) and five-category (SST-5) datasets [36]. Five-category sentiment models are benchmarked against the SST-5 dataset of over 11,000 sentences with human labels on a five-category scale. While we use the full five-category prediction functionality, the high accuracy score on the SST-2 two-category labels is an important validation that the model is directionally accurate: the model correctly classifies positive and negative sentiment. The five-category SST-5 accuracy demonstrated by the BERT TSM (specifically, the BERT-large model variant [34]) was a best known performance at the time of its publication. While some very moderate accuracy performance gains have been reported since then (see, for example [37, 38]), these models are based off of the BERT architecture used in [34]. Table 1 shows a summary of major sentiment model architectures benchmarked against the SST-2 and SST-5 datasets.

**Accuracy (%) of exemplar sentiment models on the SST benchmarks**

| Model | SST-2 | | SST-5 | |
|---|---|---|---|---|
| | All | Root | All | Root |
| VADER (Hutto 2014 [39]) | - | 72 | - | 27 |
| Avg. word vectors (Socher 2013 [36]) | 85.1 | 80.1 | 73.3 | 32.7 |
| LSTM (Tai 2015 [40]) | - | 84.9 | - | 46.4 |
| BERT-large (Munikar 2019 [34]) | 94.7 | 93.1 | 84.2 | 55.5 |

Table 1. Performance of selected models showing the evolution of sentiment analysis. The Stanford Sentiment Treebank datasets have labels for the full sentences (“Root”) as well sub-sentences according to a grammatically-parsed tree structure (“All”). Some authors did not publish results on all the phrases and words in the tree and are left blank.

For each utterance (both “client” and “therapist”) this model was used to assign one of the five sentiment categories based on the maximum probability of that class. We further assign numerical scores to each of the categories as the basis of the quantitative analyses presented in this paper. We chose to center the numerical codes with a score of 0 corresponding to a “Neutral” sentiment utterance, -2 for “Very Negative” and “2” as “Very Positive”.

We note that in other sentiment analysis methodologies, for example VADER’s “compound score” [39], these class logit or class probability values can be used to create a continuous sentiment score for each utterance instead of the categorical (argmax) value used in this work. While such probability-averaged compound scores might offer more nuanced information about a single specific turn, the utterance-level sentiment scores will be time-averaged over all all the speaker’s turns, mitigating the impact of individual mis-identified utterances. The categorical score produces a much higher variance when the time-average is taken ($\sigma^2_{compound}$=0.056 vs. $\sigma^2_{categorical}$=0.112 session scores for the corpus used here) providing a better chance to catch extremes of sentiment at the total session level.

After the transcription, diarization, role attribution and turn-level sentiment inference, the annotated transcripts look like the example in Table 2.

| turn | start (s) | end (s) | text | speaker | sentiment | Likert |
|---|---|---|---|---|---|---|

| | | | | | | code |
|---|---|---|---|---|---|---|
| **95** | 405.99 | 411.54 | think that the conversations that we’ve had over the last few sessions have really been eye opening | client | Negative | -1 |
| **96** | 411.58 | 422.17 | Um, it’s been very interesting to reflect on, um, like assessing what I’m feeling in the moment and taking a pause to just observe. | client | Positive | 1 |
| **97** | 422.89 | 427.36 | Um, and that has actually given me some insight on what I’ve noticed. | client | Positive | 1 |
| **98** | 428.18 | 434.35 | And because I noticed those things, like I think naturally the behavior, um, I don’t end up | client | Neutral | 0 |
| **99** | 435.07 | 442.18 | reacting as quickly or as, I guess, primitive, I believe to those feelings. | client | Neutral | 0 |
| **100** | 443.40 | 452.09 | So I think that naturally there is less overwhelm because of that intention to observe for sure. | client | Negative | -1 |
| **101** | 452.93 | 453.77 | Yeah, I love that. | therapist | Very Positive | 2 |
| **102** | 454.25 | 454.57 | I love that. | therapist | Very Positive | 2 |
| **103** | 455.39 | 461.62 | I’m so happy that you found some of the things that we’ve talked about beneficial and helpful. | therapist | Positive | 1 |
| **104** | 461.88 | 468.45 | Ah, and I absolutely think that, you know, much of the first step in a lot of these things is just recognition and awareness. | therapist | Positive | 1 |

Table 2. Example section from a diarized and sentiment-annotated transcript. The selection starts at speaking turn number 95, approximately 406 seconds into the session, to show utterances and scores from the middle of the session, where therapeutically meaningful interactions are taking place.

An example timeline of the utterance level sentiment score, with a smoothed curve, is shown in Figure 2. The time evolution and locations of peaks or high and low sentiment are a rich feature space for sentiment analysis and could be explored for psychotherapy process variables such as rupture / repair, critical incidents, or other significant events.

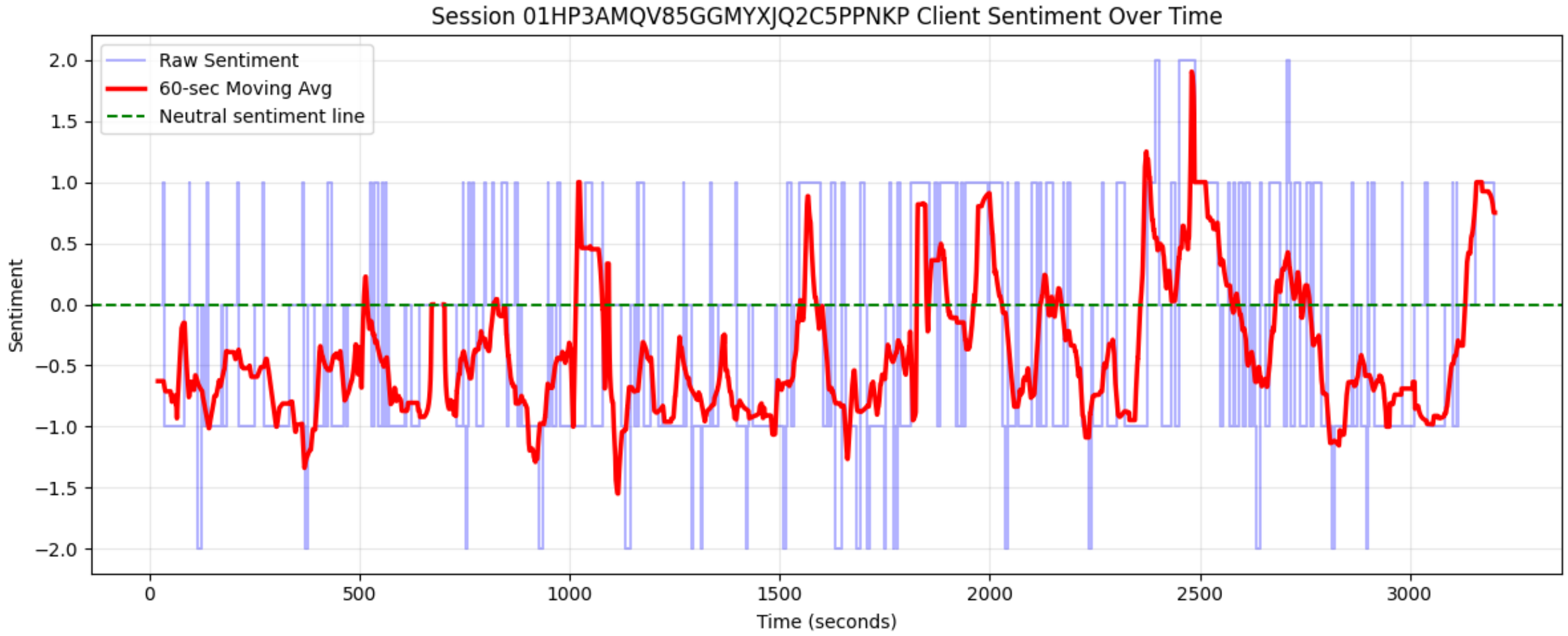

Figure 2. Example sentiment score timeline for a single session. The levels, in blue, mark the utterance level sentiment score, the smoothed curve shows a 60-second moving average of client sentiment.

### 3.2 Session Sentiment Score

Total session sentiment scores were computed by taking the time-weighted average of the utterance-level sentiment labels for that particular speaker. This feature is designed to reflect the total amount of time during the session spent expressing negative or positive sentiment. Note that this definition is always normalized to the participant's speaking time and does not directly measure the client or therapist's overall speaking time. For example, the session sentiment score for the session plotted in Fig. 2 is -0.23, an indication that, while there were some notable excursions of positive sentiment, during the majority of the session the client's language sentiment was negative.

We computed both the client and therapist sentiment scores for these sessions to investigate how much of the sentiment score might be attributed to the topics discussed in the session versus the client's internal emotional state. This is discussed further in the next session.

## 4 Results

### 4.1 Distribution of scores

By the inclusion criteria detailed above, 751 therapy sessions were transcribed and had sentiment scores computed for both the client and therapist. The overall distribution of client and therapist is described in Table 3 and Figure 3. A paired samples t-test between the client and therapist sentiment scores in each session confirms that these scores are distinct populations (t-statistic = 43, p-value 6 * 10^-206).

| | client | therapist |
|---|---|---|
| count | 751 | 751 |

| mean | -0.024 | 0.225 |
|---|---|---|
| std | 0.157 | 0.160 |
| min | -0.415 | -0.286 |
| max | 0.387 | 0.995 |

Table 3. Distribution of session-level sentiment scores for client and therapist.

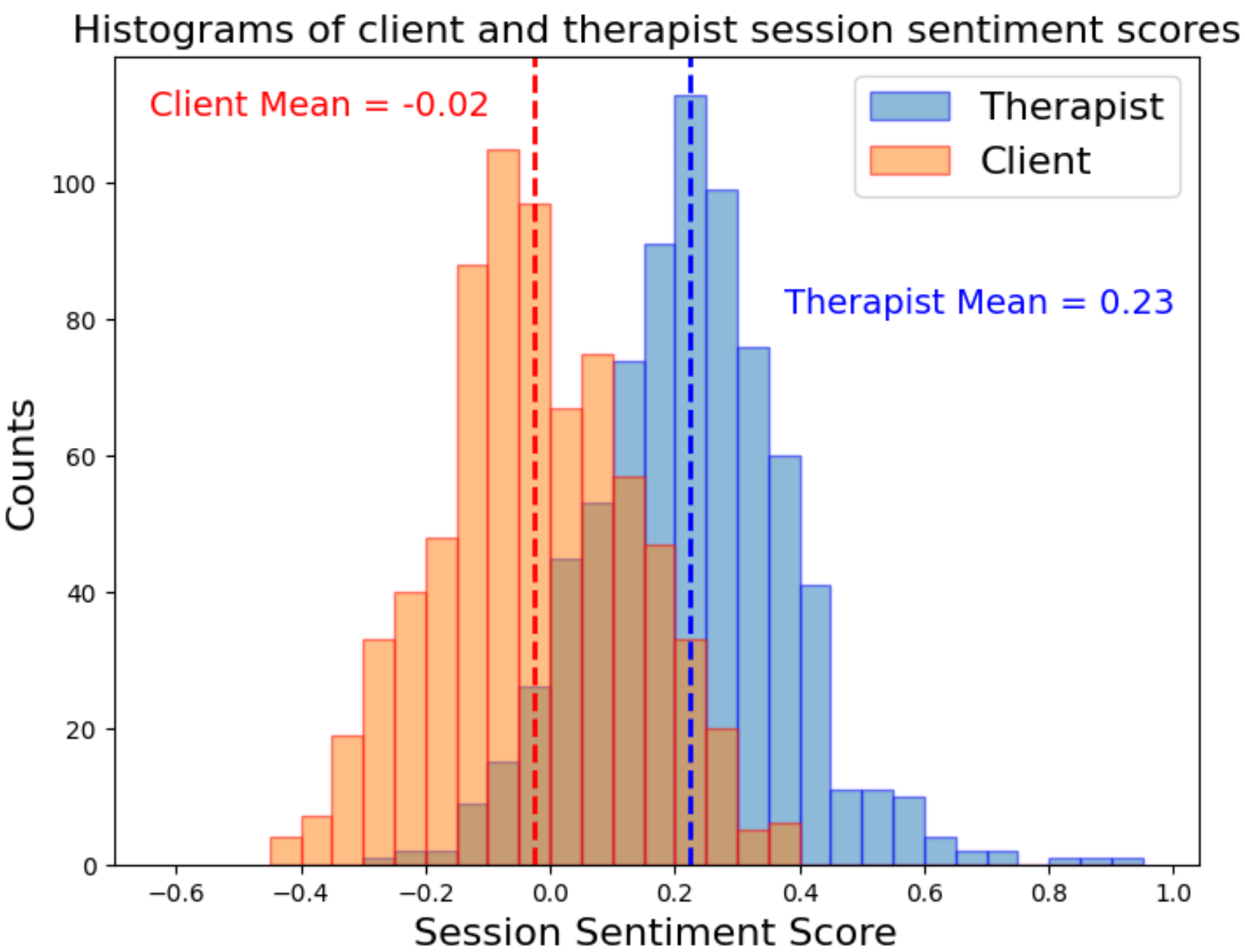

Figure 3. Histogram distribution of client and therapist sentiment scores for 751 individual (one-on-one) therapy sessions. Each distribution is annotated with its mean value.

### 4.2 OQ correlations

The Pearson correlation coefficients between the client sentiment and the overall OQ score as well as the three category subscores is shown in Table 4. In addition, the therapist sentiment score correlations were computed. This was done in order to assess the extent to which the sentiment score correlations could be driven by the negative or positive-sentiment associated topics discussed in the session versus being an intrinsic measurement of the patient's emotional state. In fact, the therapist session scores have very low/no correlation to the patient OQ subscores and the overall scores while being strongly correlated to the client sentiment score. The client sentiment scores show negative correlations to each of the OQ subscores which is expected as these scores are calibrated to be measures of discomfort or dissatisfaction (i.e. higher OQ scores are expected to be associated with lower sentiment). The overall "CurrentScore" and "Symptom Distress" subscales show moderate negative correlations while the "Social Role" and "Interpersonal Relations" subscores are weakly correlated to the client sentiment score. All of these correlations are directionally intuitive. These correlations are summarized in Table 4.

| | Social Role | Symptom Distress | CurrentScore | Interpersonal Relations | Session Sentiment | Therapist Sentiment |
|---|---|---|---|---|---|---|
| **Social Role** | 1.00 | 0.69 | 0.82 | 0.63 | -0.19 | 0.05 |
| **Symptom Distress** | 0.69 | 1.00 | 0.96 | 0.63 | -0.34 | -0.04 |
| **CurrentScore** | 0.82 | 0.96 | 1.00 | 0.80 | -0.31 | -0.02 |
| **Interpersonal Relations** | 0.63 | 0.63 | 0.80 | 1.00 | -0.20 | -0.01 |
| **Session Sentiment** | -0.19 | -0.34 | -0.31 | -0.20 | 1.00 | 0.50 |
| **Therapist Sentiment** | 0.05 | -0.04 | -0.02 | -0.01 | 0.50 | 1.00 |

Table 4: Pearson correlations between OQ-45 subscores and session sentiment for client (the "Session Sentiment" column) and therapist.

While the individual 45 OQ questions solicit verbal responses, they are ordinal and the correlations between them and the sentiment session scores can be computed using the same linear (0,...,4) numerical mapping that produces the overall and subscale scores. The Pearson correlation coefficients between 45 individual questions and client session sentiment score fall between -0.25 and 0.23 with a mean of -0.08. This indicates that the stronger subscale-level correlations are due to an aggregated measurement across many questions in the instrument.

We take the correlations between overall score, and symptom distress OQ categories as a validation that this session level sentiment feature provides therapeutically useful information from a psychotherapy session which could be used in the absence of other instruments. We interpret the comparatively absent correlation to the therapist session sentiment score versus the client session sentiment scores as evidence that the transcript sentiment scores are a reflection and measurement of the individual's emotional state at that time.

An important note regarding the interpretation of these correlations concerns the (non-)independence of the sessions nested within clients. Each client should have a unique relationship between their self-reported OQ scores and their spontaneously-expressed verbal sentiment and the clusters of client session numbers range from 2 to 32 in this session data. The possible directional impact of this on our correlations is not obvious a priori, but strong intra-client similarly would make effective sample size smaller than the nominal sample size. We can show that the size and directionality of the correlations presented in Table 4 persist for the subset of clients with 5 (the median of the count distribution) or fewer sessions. Restricting the calculation to a subset with small session counts limits the impact of within-cluster effects for clients. We see minimal impact on the correlations between session sentiment and OQ-45 subscales. These comparisons are shown in Table 5.

| | Session Sentiment (All Sessions [n=751]) | Session Sentiment (< median sessions per client [n=91]) |
|---|---|---|
| **Social Role** | -0.19 | -0.17 |

| Symptom Distress | -0.34 | -0.32 |
|---|---|---|
| CurrentScore | -0.31 | -0.32 |
| Interpersonal Relations | -0.20 | -0.33 |
| Therapist Sentiment | 0.50 | 0.49 |

Table 5: Pearson correlations between OQ-45 subscales and session sentiment for the full corpus of sessions (n=751) and the subset of individuals with a small (median or fewer, n=91) set of session to limit any effect of large, possibly non-independent, clusters of individual client sessions.

### 4.3 Progress and Alert Indicators

The OQ-45 additionally produces alert indicators based on the value and evolution of the OQ scores during the course of therapy. These codes aim to show whether the patient is functioning in the normal range, making an (in-)adequate rate of progress, or at risk of negative outcomes or dropping out of care. These are computed by both a "Rational Method" using clinically-derived algorithms and an "Empirical Method" which uses statistically generated expected recovery curves and a summary of the coding are as follows.

- White alert: The client is functioning in the normal range.
- Green alert: The client is making an adequate rate of change, and no change in the treatment plan is recommended.
- Yellow alert: The client's rate of change is not adequate.
- Red alert: The client is not making the expected level of progress and is at risk of premature dropout or a negative outcome.

In Table 5 we show the results of one-sided ANOVA calculation on the patient sentiment score for both the Rational and Empirical progress code. The ANOVA was calculated with a Type II sum of squares since the class counts for both the Rational and Empirical models were unbalanced. Both the patient and therapist scores have different means per categories in both models, however the F statistic is higher for the client sentiment score.

| feature | OQ model | Sum of Squares | df | F-statistic | P(>F) |
|---|---|---|---|---|---|
| Client Sentiment | Rational | 0.807 | 3 | 12.71 | 7.7x10^-8 |
| | Empirical | 0.743 | 3 | 11.66 | 3.0x10^-7 |
| Therapist Sentiment | Rational | 0.319 | 3 | 3.56 | 0.014 |
| | Empirical | 0.474 | 3 | 5.44 | 0.001 |

Table 5. ANOVA calculations for sentiment features and Rational/Empirical OQ alert codes.
The distributions per category are shown in Fig 4. Notably, the category with the lowest mean client sentiment score is “Red: high risk” for the Empirical model, as expected, but “Yellow” for the Rational model.

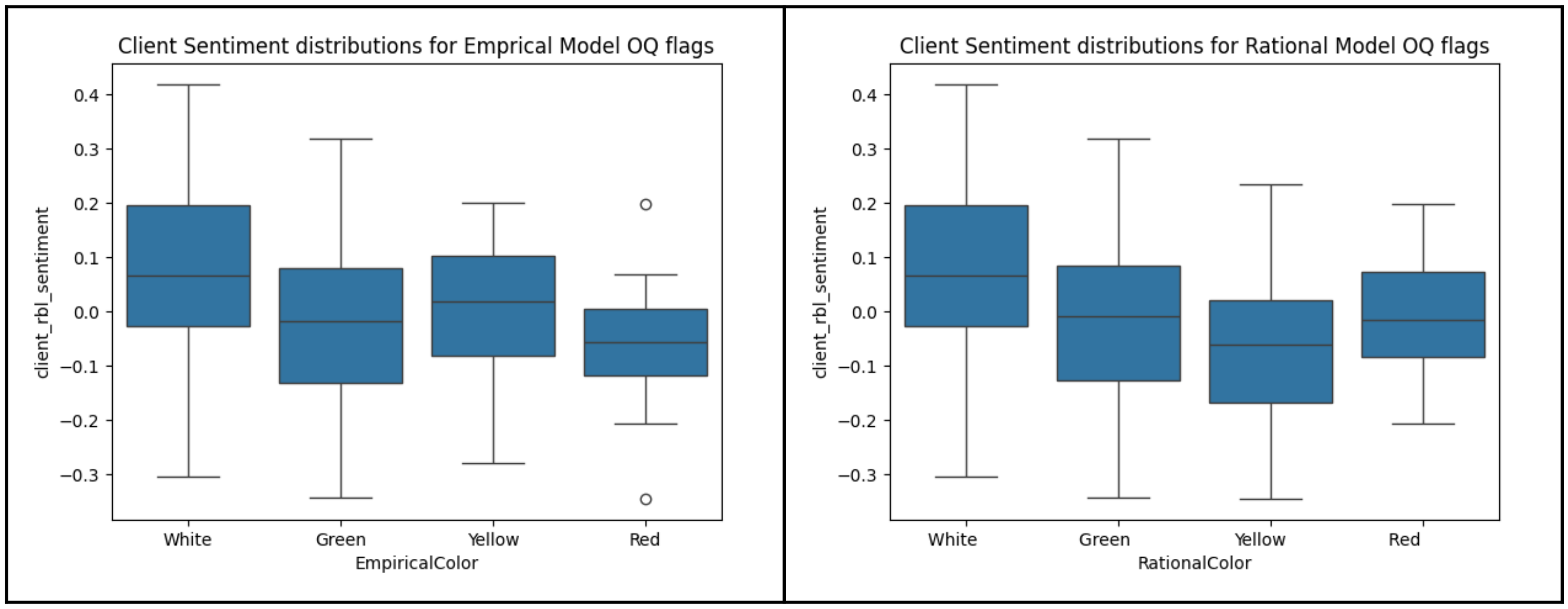

Figure 4. Boxplots by OQ-45 alert indicator category.

The observation that the sentiment features can statistically discriminate between OQ-45 alert categories suggests that this metric may offer practical utility in tracking patient progress.

## 5 Discussion

### 5.1 Summary

The aim of the present study was to characterize a proposed session sentiment score calculated with context-aware Transformer models. This sentiment feature was designed as a time-weighted average of the positive, negative, and neutral utterances mapped to a linear numerical scale. When paired by session, the session sentiment scores for the therapist and client were observed to be statistically distinct, with the mean of the therapist sentiment score being significantly higher.

We investigated the correlations between the session sentiment scores and the client’s OQ-45 survey overall and category scores. The client session sentiment score has moderate correlation to the overall OQ score ($r = -0.31$) and the “Symptom Distress” subscore ($r = -0.34$), both of which are in the expected direction. The correlations with “Social Role” ($r = -0.19$) and “Interpersonal Relations” ($r = -0.20$) OQ subscores are weak, but also in the expected direction.

Further, analysis of variance tests determine that there are significant differences between the means of the client sentiment scores for clients in different progress alert categories according to both the OQ “Rational” and “Empirical” models. These investigations show that this sentiment approach may have utility as an instrument to investigate symptom distress and progress in the clients’ course of care. This approach could provide valuable information to supplement existing self-report survey instruments because it does not require a separately administered instrument and the data reflect revealed information instead of self-reported information.

The present findings open several promising avenues for future research that could substantially advance our understanding of sentiment analysis as a clinical and research tool in psychotherapy. While this study focused on session-level aggregated sentiment scores, the within-session variability of sentiment may itself be clinically meaningful. Research on psychological flexibility suggests that the capacity to experience and adaptively shift between emotional states is a fundamental aspect of mental health [41]. Additionally, since the therapeutic alliance is one of the most robust predictors of psychotherapy outcome [42], future research should investigate whether discrepancies between client and therapist sentiment scores might signal alliance ruptures or empathic failures. To establish sentiment analysis as a mature psychotherapy research methodology, future studies must also systematically compare these measures with established process instruments such as the Core Conflictual Relationship Theme (CCRT) method [43] or the Assimilation of Problematic Experiences Scale (APES; Stiles et al., 1990 [44]). Such convergent validity studies would help researchers understand both the unique contributions of automated sentiment analysis and the complementary information provided by traditional manual coding systems.

Given the deliberate practice supervision context of the present study, sentiment analysis may offer unique opportunities for therapist development. Future research could explore how providing trainees with visual representations of session sentiment trajectories, both their own and their clients', enhances their ability to recognize and respond to emotional states in real time. Comparing sentiment patterns between novice and expert therapists might reveal specific competencies related to emotional attunement and regulation.

### 5.2 Limitations

Several limitations should be considered when interpreting the present findings. First, the sentiment model used in this study was originally trained on general-domain text corpora rather than psychotherapy-specific language. Although the five-category BERT-based model demonstrates strong performance on benchmark datasets, certain linguistic phenomena common in psychotherapy sessions, such as metaphorical expressions of distress, clinically nuanced disclosures, or indirect communication, may not be fully captured. Second, all session transcripts originated from a single telehealth training clinic, which may limit generalizability. Trainees in supervision-driven environments often follow structured conversational patterns, and their clients may differ demographically or clinically from those in community or specialty care settings. Additional replication across broader therapist populations, treatment modalities, and clinical contexts is necessary to evaluate the robustness of the proposed sentiment metric.

A further limitation involves the role of language, culture, and speech variation. Automated transcription and sentiment models may be less accurate for clients or therapists who use non-standard dialects, speak English as an additional language, draw on culturally specific expressions. Cultural differences in emotional expression and communication style may influence how sentiment is conveyed in language, potentially leading to systematic bias in model outputs. Future work should evaluate how sentiment scores perform across diverse linguistic and cultural groups and consider model fine-tuning or adaptation to better support equity in automated psychotherapy research tools.

Finally, although the study found theoretically coherent associations between sentiment scores and OQ-45 indices, the cross-sectional correlational design limits causal inference. Because sentiment is only one component of emotional expression, it remains unclear whether the observed relationships reflect underlying distress, the content of session discussions, or unmeasured third variables such as alliance quality or contextual stressors. Explicitly longitudinal experimental designs will be needed to determine whether sentiment scores can predict changes in clinical outcomes or meaningfully inform clinical decision-making.

### 5.3 Conclusion

This study demonstrates an application of Transformer-based sentiment analysis to psychotherapy sessions using a fine-grained sentiment model trained on general domain text corpora. This proposed sentiment feature aligns in theoretically-coherent ways to the OQ-45 instrument, an established indicator of client functioning. Client sentiment scores showed moderate associations with overall OQ-45 distress and symptom distress, and weak but directionally consistent associations with interpersonal and social functioning. The session sentiment scores also showed significant differentiation across the OQ alert categories which are used to indicate lack of progress or deterioration during the course of care. These findings suggest that computational NLP tools are now sophisticated enough to provide an adjunctive indicator of clients' distress and progress. Such computational tools are attractive because they provide low-burden, objective indicators that complement self-report measures. As well as establishing construct-validity evidence for this sentiment feature, the data pipeline which was instantiated to do this work demonstrates the feasibility of computational NLP methods in real-world settings. Specifically, this study was done with a corpus of recorded telehealth sessions taken from a variety of recording environments and with therapists of a variety of experience levels. Further, the computational tools used in this data pipeline are all open source and can be run on commodity hardware, allowing a very low-cost and privacy-preserving tool.

## 6 Author Contributions

TR and DF conceived the research question and proposed the fine-grained sentiment feature. DF created the data pipeline and performed the statistical analysis. PA and TR coordinated the matching of clinic data to transcripts. PA and DF performed the hand labeling and quality assurance of transcript data. TR and AV conducted the literature search and revised the initial manuscript. All authors reviewed and approved the final manuscript.

## 7 Acknowledgements

The authors would like to acknowledge David Montgomery for his early feedback on the labeling workflow and Ben Fineman for his assistance coordinating with the OQ and video platforms.

## 8 Ethics declarations

### 8.1 Competing interests

The authors declare that the research was conducted in the absence of any commercial or financial relationships that could be construed as a potential conflict of interest.

### 8.2 IRB/Ethics statement

This study received IRB approval #25-140-1012 through the Biomedical Research Alliance of New York (BRANY).

### 8.3 Consent to Participate

Every human participant (therapists and clients) have provided their informed consent to participate in this research.

## 9 Funding Declaration

This research received no funding.

## 10 Data Availability Statement

The recording and transcript data are retained on a HIPAA compliant storage and are not publicly available as per the consent agreement and general HIPAA guidelines. The anonymized summary data used to perform the statistical analyses are available upon reasonable request.